\pdfoutput=1
\documentclass{article}

\PassOptionsToPackage{numbers, compress}{natbib}

\usepackage[preprint]{neurips_2024}

\usepackage[utf8]{inputenc} 
\usepackage[T1]{fontenc}    
\usepackage{hyperref}       
\usepackage{url}            
\usepackage{booktabs}       
\usepackage{amsfonts}       
\usepackage{nicefrac}       
\usepackage{microtype}      
\usepackage{xcolor}         
\usepackage{subcaption}
\usepackage{caption}
\usepackage{geometry}
\usepackage[pdftex]{graphicx}
\usepackage{epstopdf}
\usepackage{tikz}
\usepackage{colortbl}
\usepackage{amsmath}
\usepackage{multirow}
\usepackage{colortbl}
\usepackage{xcolor}
\usepackage[most]{tcolorbox}
\usepackage{array}
\usepackage{algorithm}
\usepackage{algorithmic}
\usepackage{amsmath}
\usepackage{listings}
\usepackage{xcolor}
\usepackage{amsmath}

\lstdefinestyle{overleaf}{
    backgroundcolor=\color[rgb]{0.95,0.95,0.92},   
    commentstyle=\color[rgb]{0,0.6,0},
    keywordstyle=\color{magenta},
    numberstyle=\tiny\color[rgb]{0.5,0.5,0.5},
    stringstyle=\color[rgb]{0.58,0,0.82},
    basicstyle=\ttfamily\footnotesize,
    breakatwhitespace=false,         
    breaklines=true,                 
    captionpos=b,                    
    keepspaces=true,                 
    numbers=left,                    
    numbersep=5pt,                  
    showspaces=false,                
    showstringspaces=false,
    showtabs=false,                  
    tabsize=2
}
\lstdefinestyle{mocov3}{
  backgroundcolor=\color{white},
  basicstyle=\fontsize{7.5pt}{7.5pt}\ttfamily\selectfont,
  columns=fullflexible,
  breaklines=true,
  captionpos=b,
  commentstyle=\fontsize{7.5pt}{7.5pt}\color[rgb]{0.25,0.5,0.5},
  keywordstyle=\fontsize{7.5pt}{7.5pt}\color[rgb]{0.85,0.18,0.50},
}
\title{Freeplane: Unlocking Free Lunch in Triplane-Based Sparse-View Reconstruction Models}

\author{Wenqiang Sun$^{1,3}$,\quad Zhengyi Wang$^{2,3}$,\quad Shuo Chen$^{2}$,\quad 
Yikai Wang$^{2}$,\quad Zilong Chen$^{2,3}$, \\ \textbf{Jun Zhu}$^\dag$$^{2,3}$\textbf{,}\quad \textbf{Jun Zhang}$^\dag$$^{1}$
\\$^1$Department of Electronic and Computer Engineering, HKUST \\
$^2$Department of Computer Science and Technology, Tsinghua University \quad $^3$ShengShu  \\
\texttt{wsunap@connect.ust.hk}; \\
\texttt{\{wang-zy21, chenshuo20, chenz122\}@mails.tsinghua.edu.cn}; \\
\texttt{yikaiw@outlook.com};~
\texttt{dcszj@tsinghua.edu.cn};~
\texttt{eejzhang@ust.hk};
}

\begin{document}

\maketitle

\let\thefootnote\relax\footnotetext{$^\dag$Corresponding author}
\begin{abstract}
Creating 3D assets from single-view images is a complex task that demands a deep understanding of the world. Recently, feed-forward 3D generative models have made significant progress by training large reconstruction models on extensive 3D datasets, with triplanes being the preferred 3D geometry representation. However, effectively utilizing the geometric priors of triplanes, while minimizing artifacts caused by generated inconsistent multi-view images, remains a challenge. In this work, we present \textbf{Fre}quency modulat\textbf{e}d tri\textbf{plane} (\textbf{Freeplane}), a simple yet effective method to improve the generation quality of feed-forward models without additional training. 
We first analyze the role of triplanes in feed-forward methods and find that the inconsistent multi-view images introduce high-frequency artifacts on triplanes, leading to low-quality 3D meshes. Based on this observation, we propose strategically filtering triplane features and combining triplanes before and after filtering to produce high-quality textured meshes. These techniques incur no additional cost and can be seamlessly integrated into pre-trained feed-forward models to enhance their robustness against the inconsistency of generated multi-view images. 
Both qualitative and quantitative results demonstrate that our method improves the performance of feed-forward models by simply modulating triplanes. All you need is to modulate the triplanes during inference.
\end{abstract}


\section{Introduction}    \label{intro}

Recently, generative models have showcased substantial progress in generating realistic images and videos through the integration of extensive datasets and efficient architectures. When it comes to 3D generation, however, the limited datasets constrain generative models from producing high-quality 3D assets that can be used for industrial manufacturing and artistic creation. To resolve this issue, DreamFusion and subsequent works \cite{poole2022dreamfusion,lin2023magic3d,wang2024prolificdreamer} have tried to distill 2D image prior from Stable Diffusion \cite{rombach2021high} into 3D scene representation using the Score Distillation Sampling (SDS) technique. Nevertheless, the SDS-based per-scene optimization encounters challenges of 3D inconsistency and time-consuming computations, making it impractical for real-world scenarios.

To achieve fast and generalizable 3D generative models, early studies \cite{gao2022get3d,chan2022efficient} develop the triplane-based 3D GAN architecture, benefiting from the computation-efficient and expressive nature of triplanes. Despite generating high-quality 3D assets, the generalization capacity of these models is constrained by their unstable training process and non-scalable architectures. Inspired by the achievements of the transformer architecture \cite{vaswani2017attention,ouyang2022training} in natural language processing, LRM \cite{hong2023lrm} introduces a transformer-based reconstruction model to reconstruct high-quality 3D meshes from a single image. Equipped with the scalable architecture and large-scale 3D datasets, LRM has become a promising backbone for 3D generation. Built upon LRM, Instant3D \cite{li2023instant3d} and InstantMesh \cite{xu2024instantmesh} introduce a multi-view diffusion model to achieve a fast single image to 3D generation process.  
To fully utilize the 3D geometry prior in triplanes, CRM \cite{wang2024crm} suggests generating highly detailed triplanes from multi-view images using a UNet-based convolutional network. Leveraging high-resolution triplanes as strong priors, CRM generates textured meshes with detailed geometry. Nevertheless, CRM fails to achieve satisfactory 3D generation because high-resolution triplanes are highly sensitive to inconsistencies in generated multi-view images. 

These feed-forward methods, consisting of a multi-view diffusion model and a sparse-view reconstruction model, encounter a common challenge: the quality of generated meshes is inevitably influenced by the 3D inconsistency of multi-view images. Existing works mainly focus on training more powerful multi-view diffusion models and scalable reconstruction models, while the internal properties of triplanes remain largely under-explored. In light of this, we raise the question for 3D generation: \textbf{is it possible to alleviate the artifacts caused by inconsistent multi-view images directly within the triplanes?} 

\begin{figure}[t]
    \centering
    \includegraphics[width=\linewidth]{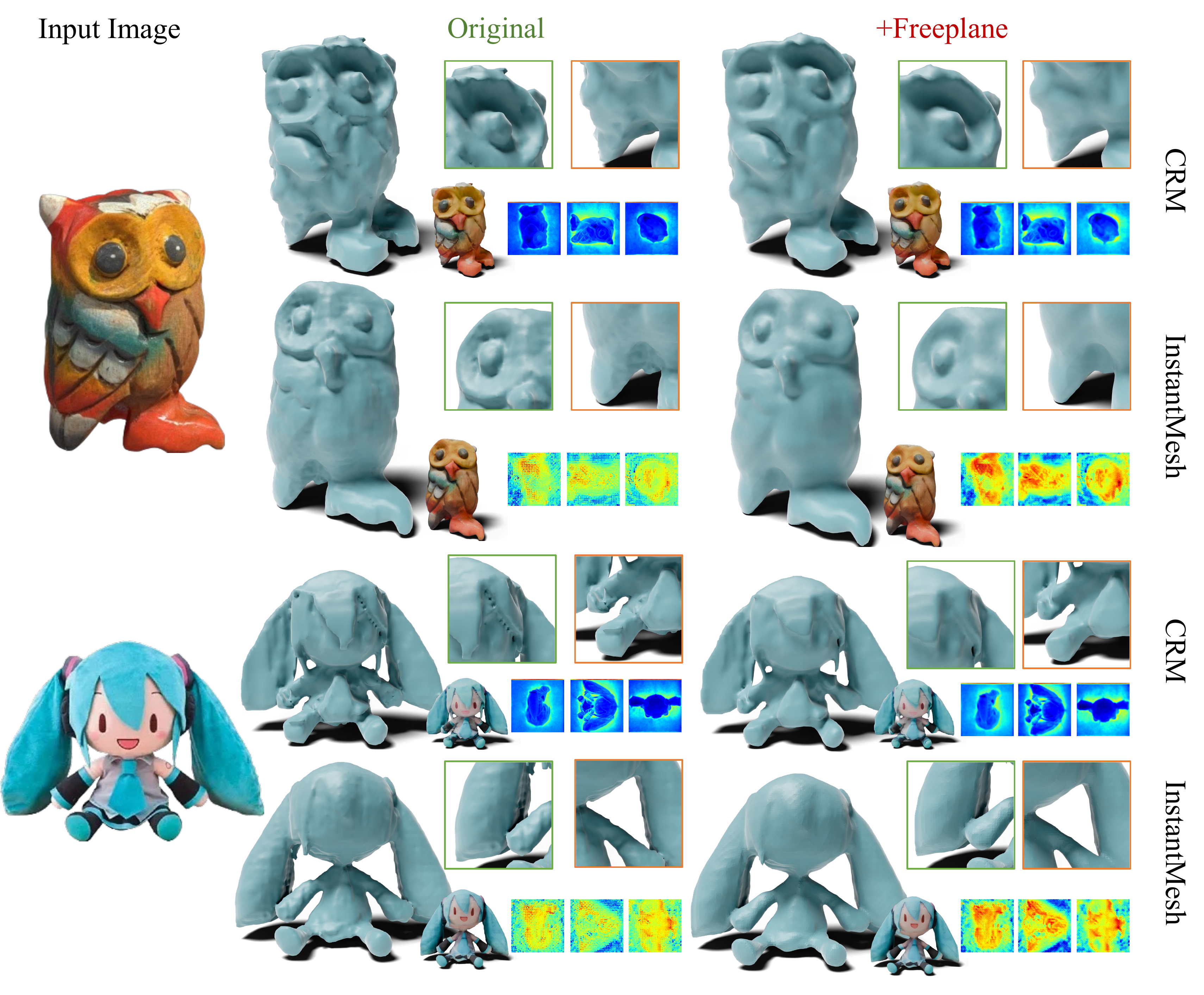}
    \caption{We present \textbf{Freeplane}, a method that substantially refines the mesh quality of feed-forward generative models without additional costs: no training or fine-tuning, no extra memory required, and only a few lines of code.}
    \label{fig:head}
\end{figure}

In this work, we propose a simple yet effective method that surprisingly addresses the aforementioned challenge with just a few additional lines of code. Specifically, we examine the role of triplanes in feed-forward models and find that inconsistent multi-view images inherently introduce high-frequency artifacts into the triplanes, resulting in rough and uneven meshes. Drawing on our analysis, we propose two strategies to improve the 3D generation quality without additional cost. One is frequency modulation, applied to the triplanes, which reduces the negative effects caused by inconsistent multi-view images. The other is the combination of triplanes before and after filtering to produce detailed textured meshes, which utilizes the former to predict the SDF values and the latter to acquire the texture. We refer to our approach as \textbf{Fre}quency modulat\textbf{e}d tri\textbf{plane} (\textbf{Freeplane}), which is free of additional training and computational overhead. 

Our Freeplane framework can be seamlessly integrated with existing feed-forward methods, improving the generation quality within a few lines of code. We conduct a comprehensive evaluation of our approach, employing CRM \cite{wang2024crm} and InstantMesh \cite{xu2024instantmesh} (two state-of-the-art open-sourced feed-forward generative models) as the base models. By incorporating Freeplane during the testing phase, these models show a noticeable enhancement in the smoothness and geometric details of generated meshes. The visualization results depicted in Fig. \ref{fig:head} confirm the effectiveness of Freeplane in reducing the artifacts arising from inconsistent multi-view images. Our contributions are summarized as follows:
\begin{itemize}
    \item We reveal the role of triplanes in feed-forward models and uncover the performance degradation caused by inconsistent multi-view images, which is further verified by empirical studies and addressed by our approach.
    \item We introduce a simple yet effective method, denoted as \textbf{Freeplane}, which fully utilizes the geometry priors in triplanes while alleviating the artifacts caused by inconsistent multi-view images. It broadly improves the smoothness and geometry details of generated meshes without requiring additional training.
    \item The proposed approach can be seamlessly integrated into existing triplane-based sparse-view reconstruction models. Both qualitative and quantitative experimental results demonstrate that our approach exhibits apparent generation quality enhancement across different methods, showing the robustness of Freeplane.
\end{itemize}

\section{Related Work}    \label{related_work}

\subsection{Neural Fields} 
The neural radiance field (NeRF) \cite{mildenhall2021nerf} is a popular representation in the 3D vision area, which uses deep neural networks to represent 3D scenes as continuous functions. Following works \cite{barron2022mip,tancik2022block,niemeyer2022regnerf,park2021hypernerf} extend the neural fields to large-scale scenes and sparse-input reconstruction tasks. However, it takes a long time to train these NeRF-based models, usually from hours to days. To accelerate the training process while keeping a low memory cost, TensoRF \cite{chen2022tensorf} proposes to formulate the radiance field as a 4D tensor and factorize the 4D tensor into low-rank components. 
In light of this, LRM chooses the triplanes as the concise and scalable 3D representation to build a generalized reconstruction model.

Although NeRF can be transformed into meshes using Marching Cubes \cite{lorensen1998marching}, the quality of extracted meshes is not assured. Previous methods aim to convert the neural fields into the implicit surface representation, such as the Signed Distance Function (SDF) \cite{yariv2021volume,wang2021neus,li2023neuralangelo}, to produce smooth and detailed meshes. Moreover, some proposed differentiable marching cubes (DiffMC) techniques \cite{remelli2020meshsdf,wei2023neumanifold} improve the mesh reconstruction quality and speed. Recently, to present the high-fidelity geometric details, Flexicube \cite{shen2023flexible} proposes to reconstruct the mesh from features of grids by dual marching cube \cite{schaefer2004dual}. The grid features include the SDF values, weights and deformation. In addition, the textures are queried from the surface. To produce high-resolution textured meshes, recent feed-forward models \cite{wang2024crm,xu2024instantmesh} adopt Flexicubes as the geometry representation.


\subsection{3D Generation} Generative models, such as Generative Adversarial Networks (GANs) \cite{goodfellow2014generative} and Diffusion Models \cite{rombach2021high}, have made remarkable progress in creating realistic and diverse images and videos. In the context of 3D generation, some early attempts directly utilize 3D assets or multi-view datasets to train 3D generative models with GANs \cite{wu2016learning,chan2022efficient,gao2022get3d} and diffusion models \cite{luo2021diffusion,nichol2022point,jun2023shap,gupta20233dgen,li2023diffusion,mo2024dit}. Despite achieving relatively satisfactory 3D shape generation, their progress is hindered by the scale, quality, and diversity of 3D datasets.  DreamFusion \cite{poole2022dreamfusion} proposes leveraging the pre-trained image diffusion models \cite{rombach2021high,saharia2022photorealistic} to optimize the 3D models with a technique called Score Distillation Sampling (SDS). Follow-up approaches aim to achieve faster optimization or more high-quality generation \cite{lin2023magic3d,wang2023score,chen2023fantasia3d,wang2024prolificdreamer,tang2023dreamgaussian,chen2023text,zhao2024flexidreamer}. 
However, these SDS optimization-based methods rely on the per-scene optimization, which are usually time-consuming, taking from minutes to hours to generate a single 3D object. In addition, Zero123 \cite{liu2023zero} introduces the view-conditioned diffusion models to enhance the 3D consistency of SDS-based optimization. To improve the multi-view consistency, the following works \cite{shi2023mvdream,wang2023imagedream,liu2023syncdreamer,long2023wonder3d} propose generating the multi-view images simultaneously.
Meanwhile, some studies \cite{liu2024one,liu2023syncdreamer,long2023wonder3d} try to acquire sparse-view images from a single image and then optimize the 3D assets based on the sparse-view reconstruction technique. However, a primal issue is that these methods require test-time reconstruction, potentially resulting in additional time consumption and degraded quality.

More recently, pioneering research has attempted to generate high-quality 3D objects using a feed-forward model  \cite{hong2023lrm,li2023instant3d,zou2023triplane,tang2024lgm,wang2024crm,xu2024instantmesh,wei2024meshlrm,zhang2024gs}, demonstrating stronger generalization capabilities and faster generation speeds compared to earlier methods. LRM \cite{hong2023lrm} firstly adopts the transformer-based architecture to achieve high-quality and fast 3D object creation. Instant3D \cite{li2023instant3d} combines a multi-view diffusion model with the transformer-based reconstruction model to generate 3D meshes from a single text or image prompt. Building on the LRM series \cite{hong2023lrm,li2023instant3d}, InstantMesh \cite{xu2024instantmesh} integrates Flexicubes into the transformer-based framework to enhance the smoothness and geometric details. MeshLRM \cite{wei2024meshlrm} redesigns the transformer-based architecture and introduces more efficient training strategies. Additionally, a differentiable marching cube \cite{wei2023neumanifold} is adopted to improve the rendering speed and quality. Instead of employing the transformer-based architecture, CRM \cite{wang2024crm} suggests generating high-resolution triplanes from orthographic multi-view images with the UNet backbone and utilizing Flexicubes \cite{shen2023flexible} as the geometry representation. Taking into account the fast rendering speed and explicit representation, some studies \cite{zou2023triplane,tang2024lgm,xu2024grm,wang2024vidu4d,zhang2024gs} adopt Gaussian Splatting \cite{kerbl20233d} as the 3D geometry representation to facilitate a fast training process. Specifically, GS-LRM \cite{zhang2024gs} extends the architecture to the scene generation task. Overall, our proposed approach can be adapted to existing triplane-based feed-forward models.

\begin{figure}[t]
    \centering
    \includegraphics[width=\linewidth]{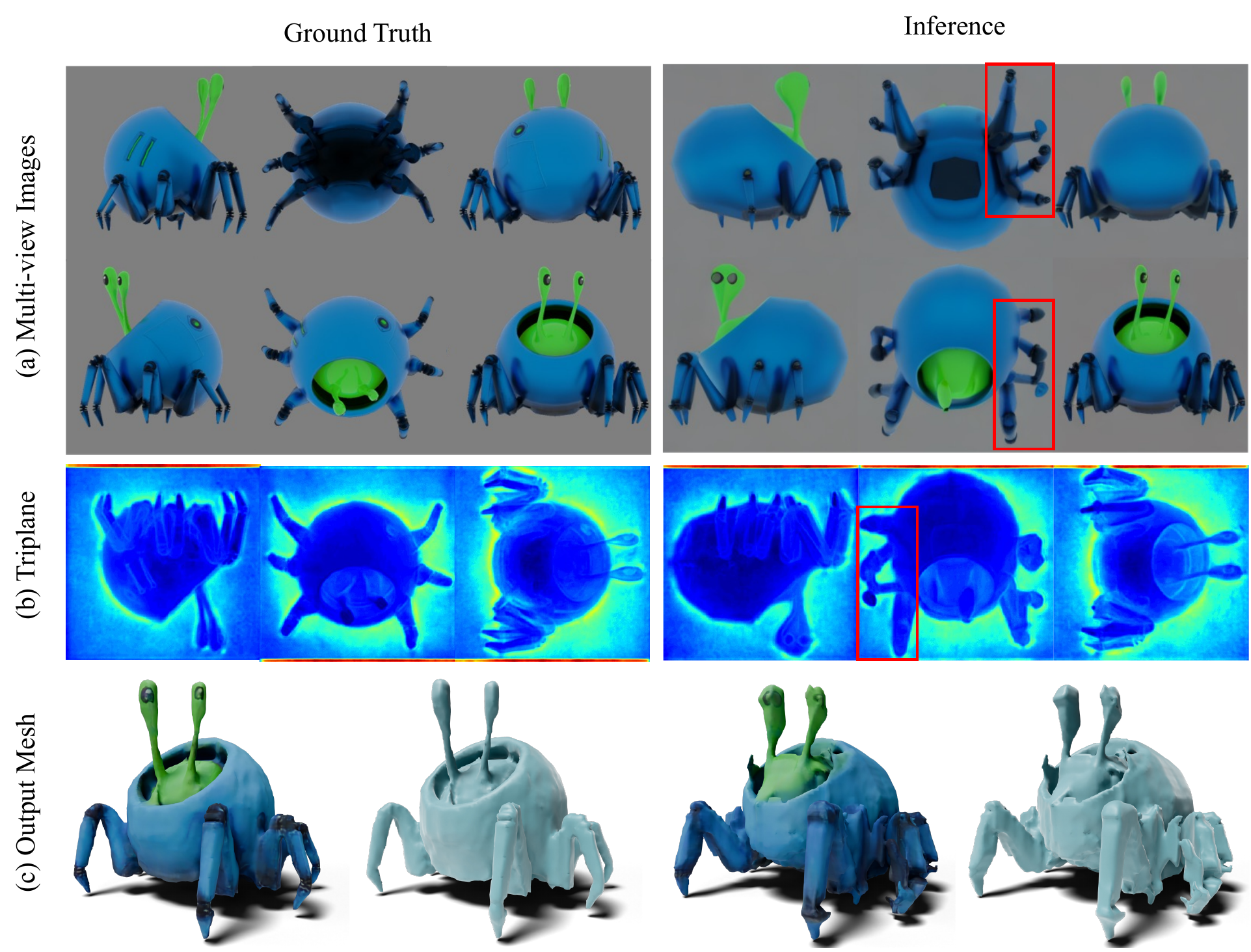}
    \caption{Our key observation is that inconsistent multi-view images cause high-frequency artifacts on the triplanes, resulting in low-quality meshes. 
    Given ground-truth images, CRM can generate a smooth and highly detailed mesh. However, during the inference stage, using the inconsistent multi-view images from diffusion models causes apparent artifacts on the triplanes, leading to low-quality 3D assets.}
    \label{fig:motivation}
\end{figure}

\section{Methodology}   \label{method}
\begin{figure}[t]
    \centering
    \includegraphics[width=\linewidth]{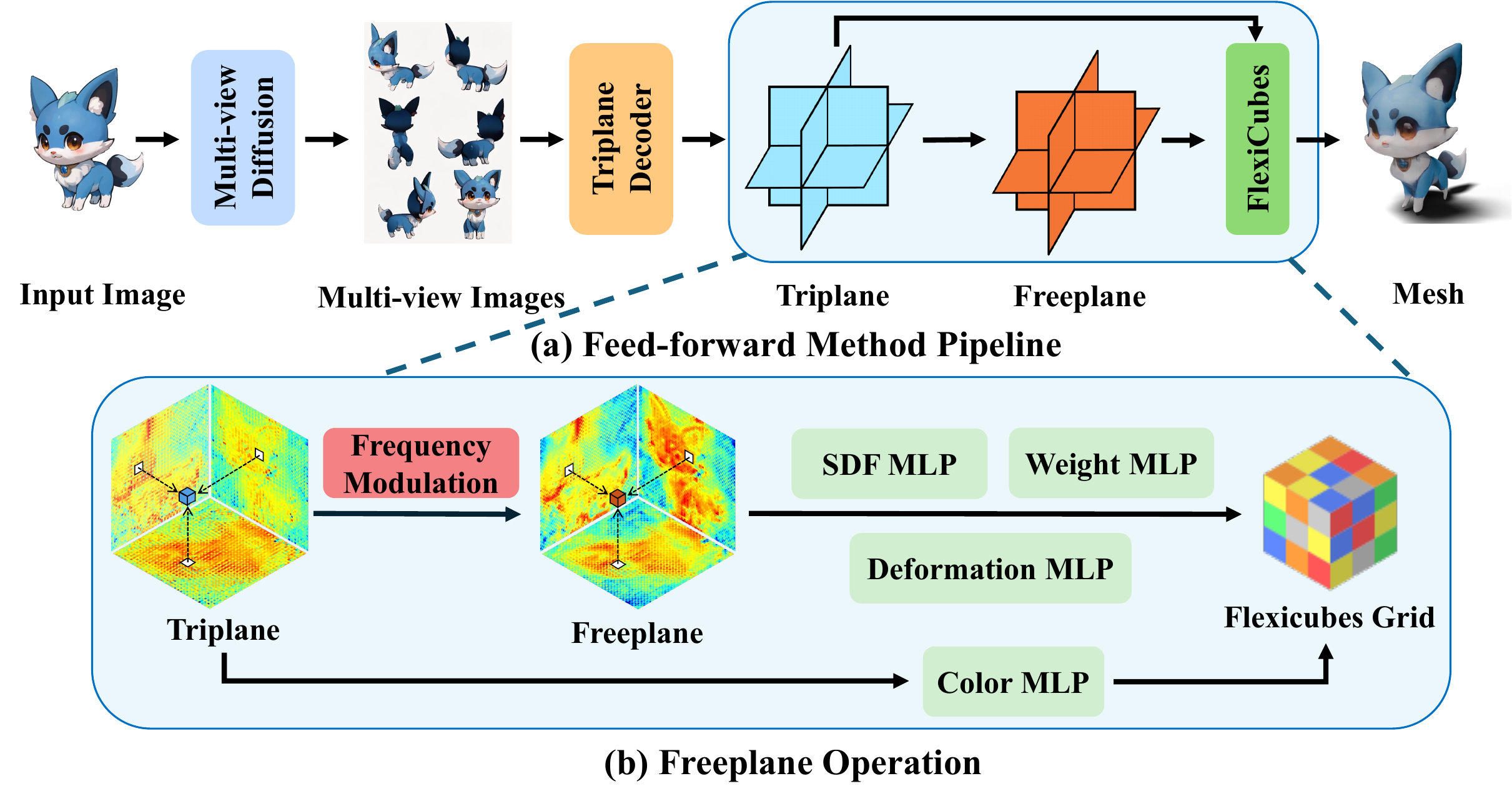}
    \caption{\textbf{Freeplane Framework.} \textbf{(a) Feed-forward Pipeline.} A single image is input into the multi-view diffusion model to generate six-view images, which are then fed into the triplane decoder. By querying the triplane features, Flexicubes are extracted to produce the textured mesh. We adopt the Freeplane approach on the triplanes. \textbf{(b) Freeplane Operation.} Low-frequency filtering is applied to modulate the original triplanes. Triplanes before filtering are used to compute texture-related features, while those after filtering are utilized to predict mesh geometry.}
    \label{fig:pipeline}
\end{figure}

Our approach is broadly adopted in triplane-based feed-forward models with a multi-view diffusion and a sparse-view reconstruction model. As shown in Fig. \ref{fig:pipeline}, the input image $I$ is sent to the multi-view diffusion model to generate multi-view images, which are then fed into the decoder to acquire the triplanes. Subsequently, triplanes with and without Freeplane are combined together to predict the geometry and texture.

In Sec. \ref{3.1.1} and \ref{3.1.2}, we provide a brief introduction about triplanes and feed-forward models. Sec. \ref{3.2} shows our key observation and analysis about triplanes. In addition, we present the details of our approach: \textbf{Freeplane}.

\subsection{Preliminaries}

\subsubsection{Triplanes} \label{3.1.1}
TensoRF \cite{chen2022tensorf} proposes to swap the original MLP used in NeRF \cite{mildenhall2021nerf} with a feature volume to accelerate the training process. It factorizes the 4D tensors into low-rank multiple vactors and matrices:
\begin{equation}
\mathcal{T}=\sum_{r=1}^{R_1} v_r^1 \circ M_r^{2,3}+\sum_{r=1}^{R_2} v_r^2 \circ M_r^{1,3}+\sum_{r=1}^{R_3} v_r^3 \circ M_r^{1,2}
\end{equation}
where $v_r^1$, $v_r^2$, $v_r^3$ are vector factors, and $M_r^{2,3}$, $M_r^{1,3}$, $M_r^{1,2}$ are matrix factors.

\subsubsection{Feed-forward Methods}   \label{3.1.2}

As mentioned before, tripalne-based feed-forward models \cite{li2023instant3d,wang2024crm,xu2024instantmesh} contain a multi-view diffusion model and a sparse-view reconstruction model. Zero123 \cite{liu2023zero} first explores the 3D prior information in pre-trained diffusion models and achieves arbitrary-view image synthesis. To further enhance the 3D consistency, the following works \cite{,shi2023mvdream,liu2023syncdreamer,shi2023zero123++,long2023wonder3d,wang2023imagedream,weng2023consistent123,lu2023direct2} try to fine-tune the pre-trained diffusion model to generate multi-view images simultaneously. The sparse-view reconstruction model aims to produce 3D meshes from the generated multi-view images, utilizing triplanes as the geometric representation. Triplanes play a significant role in linking 2D images to 3D representations, but they possess distinct properties depending on how they are acquired. Specifically, current methods treat triplanes either as 2D images or as a distinct 3D modality. Considering the generation quality and whether the model is open-source, we choose CRM \cite{wang2024crm} and InstantMesh \cite{xu2024instantmesh} as our main focuses.

Treating triplanes as 2D images, CRM adopts a variant of ImageDream \cite{wang2023imagedream} and a UNet-based reconstruction model to produce highly detailed textured meshes. Generated from the multi-diffusion model, orthographic images are fed into the UNet to acquire high-resolution triplanes. On the other hand, adhering to the design in LRM, InstantMesh regards triplanes as a novel modality and connects them with 2D images using the cross-attention mechanism. In addition, InstantMesh opts for a fine-tuned Zero123++ \cite{shi2023zero123++}, which includes multiple views from positive and negative elevations. To generate smooth meshes with geometric details, both CRM and InstantMesh adopt Flexicubes as the geometry representation.

\subsection{Free Lunch in Triplane-Based Sparse-View Reconstruction Models}     \label{3.2}

While these sparse-view reconstruction models achieve high-quality 3D generation, they encounter a gap between training and inference. It is noted that during the training phase, consistent multi-view images are used to create high-quality triplanes. However, in the inference stage, multi-view images generated by a fine-tuned diffusion model inevitably exhibit inconsistencies.
Although data perturbation methods \cite{tang2024lgm,xu2024instantmesh} are employed during training to enhance the model's robustness to inconsistencies, these augmented images fundamentally differ from those generated by the diffusion model.
As shown in Fig. \ref{fig:motivation}, the inconsistency of multi-view images causes noticeable artifacts on the triplanes, ultimately leading to low-quality meshes. Specifically, despite the global consistency of these multi-view images, there are significant conflicts and inconsistencies in certain local areas, resulting in high-frequency artifacts on the triplanes. A similar issue also appears in InstantMesh, where the inconsistency introduces high-frequency noises on the triplanes. The visualization results of InstantMesh are provided in the Appendix \ref{motivation_instant}.

Building on the aforementioned discovery, we present our simple yet effective approach, denoted as \textbf{Freeplane}. It effectively alleviates artifacts caused by inconsistent multi-view images, which is achieved by erasing the conflicts on the triplanes. Our method drastically improves the generation quality without additional costs. We use $G_M$ to represent the multi-view diffusion model, and $G_D$ to denote the triplane decoder. Triplanes from $G_D$ are denoted as $\boldsymbol{t}$, where the feature map of the $i$-th channel is $\boldsymbol{t_i}$. As shown in Fig. \ref{fig:pipeline}, the triplane features are then fed into the SDF, weight, deformation, and color MLPs to extract the Flexicube grids.

\begin{figure}[t]
    \centering
    \includegraphics[width=\linewidth]{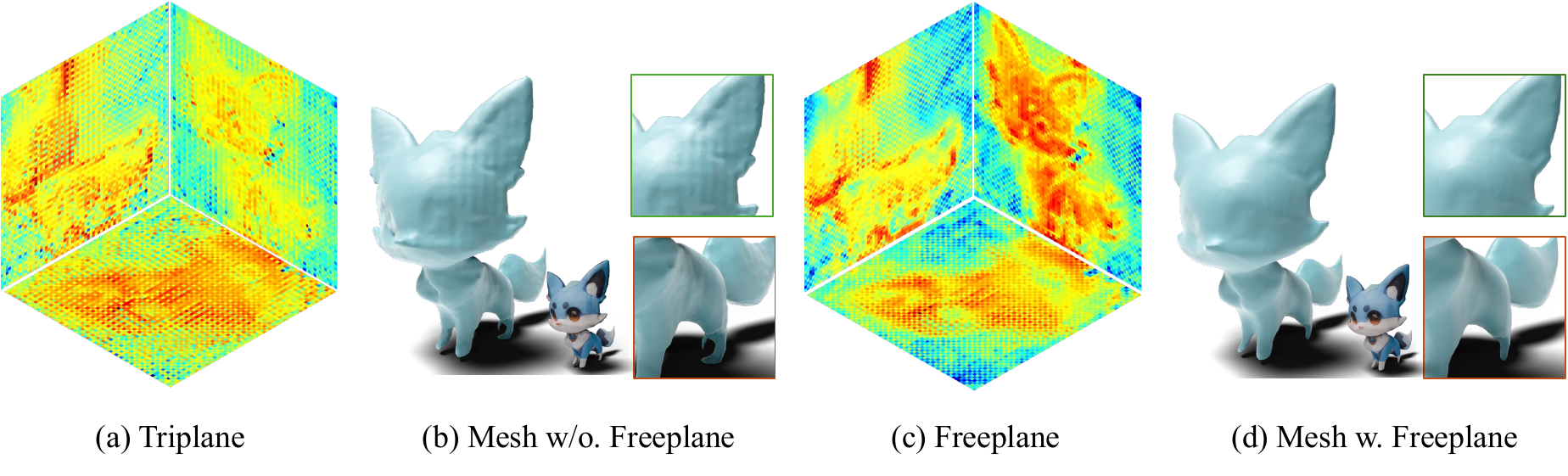}
    \caption{We present triplanes and meshes w/o. and w. \textbf{Freeplane} from InstantMesh. Our approach erases the noises and produces apparent boundaries on the triplanes, leading to a smooth surface with geometry details.}
    \label{fig:method}
\end{figure}

A key insight of our approach is that artifacts on the triplanes are essentially high-frequency components, consisting of either highly detailed conflicts or scattered noises. Consequently, a naive low-pass filter that aims to purify the high-frequency parts can be adopted to smooth the artifacts. However, we find that solely using an ordinary low-pass filter on the triplanes results in poor mesh quality. We believe that this is because the global low-frequency filtering causes a significant loss of geometric details. In our earlier analysis, we have demonstrated that the inconsistency primarily occurs in the local areas of triplanes. Building on this key observation, we propose a strategic filtering method that targets and removes high-frequency components in the surrounding areas. Moreover, to avoid the oversmoothed geometric surface, we further introduce a constraint to maintain the triplanes' boundary. Notably, our approach offers several impressive benefits. First, it erases the local artifacts on the generated triplanes to mitigate the negative influence brought by inconsistent multi-view images. Secondly, our Freeplane maintains the edge geometric information in triplanes, ensuring the apparent geometry details of 3D creations. The triplanes and 3D meshes in Fig. \ref{fig:method} confirm the effectiveness of our approach. In addition, as shown in Fig. \ref{fig:pipeline}, we propose using the triplanes before filtering to query color and the triplanes after filtering to predict geometry, ensuring fully utilize the texture priors in the pre-trained model. 

Concretely, we adopt local frequency modulation for each channel in the triplanes and concate them together. Mathematically, our approach is presented as follows:
\begin{equation}
\begin{aligned}
\boldsymbol{t}_{i,m} = \sum_{m\in {t_i}} \mathcal{K}_{l, \sigma}(|m-n|)\boldsymbol{t}_{i,n},  \\
\end{aligned}
\end{equation}
where $\boldsymbol{t}_{i,m}, \boldsymbol{t}_{i,n}$ are the values of pixel $m, n$ in the $i$-th channel of triplanes $\boldsymbol{t}$, and $\mathcal{K}_{l}$ is a low-pass kernel function to compute the local average values. In addition, the kernel size $\sigma$ of $\mathcal{K}_{l}$ determines the range of neighboring pixels involved in the computation. To preserve the edge geometric details, we introduce another low-pass kernel filter, denoted as $\mathcal{K}_{r}$:
\begin{equation}
\begin{aligned}
\boldsymbol{t}_{i,m} = \sum_{m\in {t_i}} \mathcal{K}_{l, \sigma}(|m-n|) \mathcal{K}_{r, \sigma}(|\boldsymbol{t}_{i,m}-\boldsymbol{t}_{i,n} |) \boldsymbol{t}_{i,n}.   \\
\end{aligned}
\end{equation}
For the practical implementation, we utilize the Gaussian kernel as the kernel function and use the Bilateral Filter \cite{tomasi1998bilateral} package in pytorch.

\begin{figure}[tb]
    \centering
    \includegraphics[width=\linewidth]{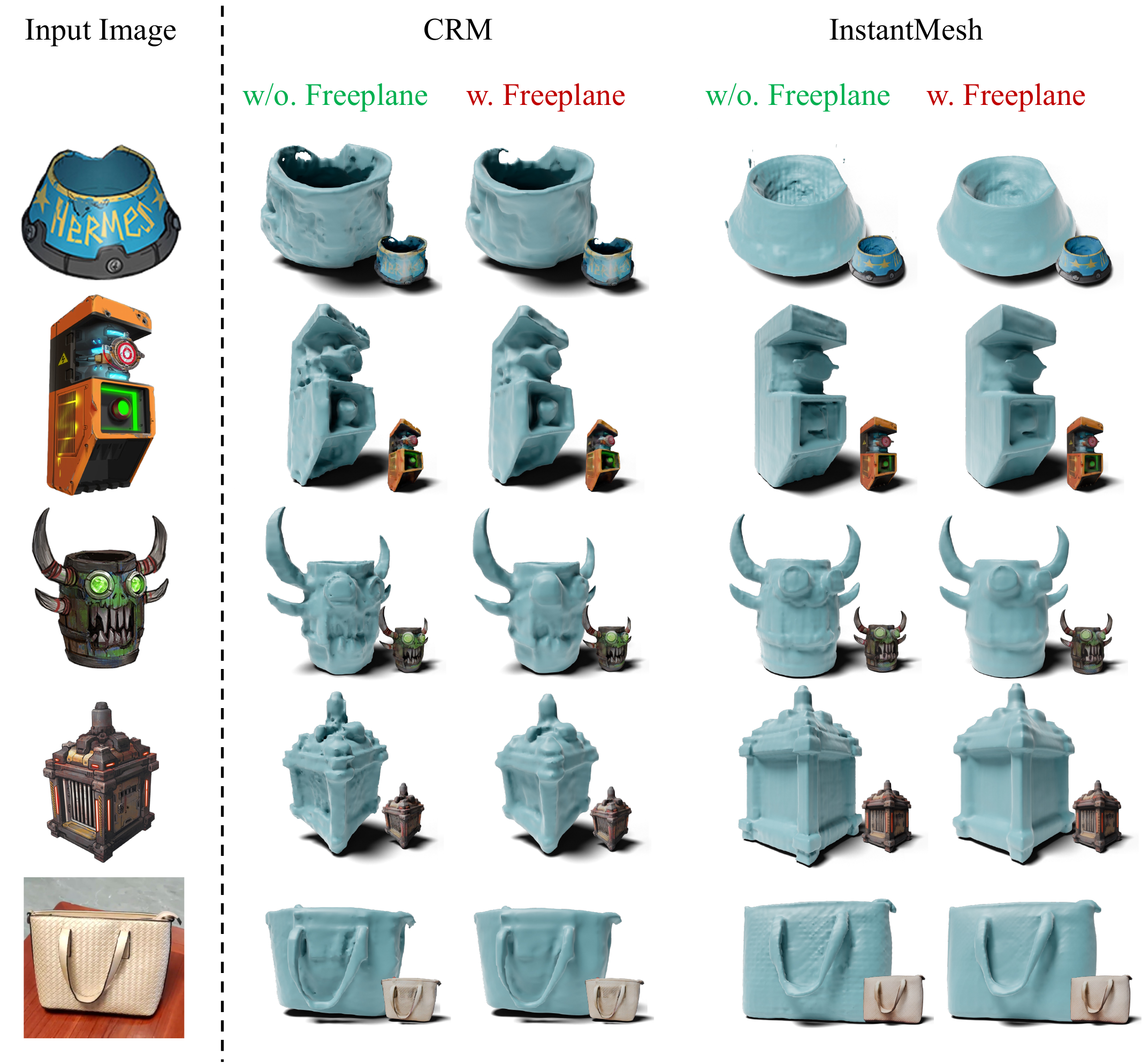}
    \caption{Qualitative Results: With Freeplane, both CRM and InstantMesh are able to generate smoother meshes with fewer artifacts while preserving their textures.}
    \label{fig:demo}
\end{figure}

\section{Experiments}    \label{exp}

\subsection{Implementation details}
We choose the open-sourced CRM and InstantMesh as our base models, which are both first-class 3D feed-forward generative models. They are pre-trained on the extensive 3D dataset Objaverse \cite{deitke2023objaverse}, providing them with robust generalization capabilities. During the inference stage, we freeze the entire model and only adjust the triplanes generated by the triplane decoder. Additionally, we apply a bilateral filter, setting the kernel size to 9 in CRM and 3 in InstantMesh. As shown in Fig. \ref{fig:ablation_left}, we provide the kernel used in our experiments. Since the bilateral filter needs to consider edges and is not a linear kernel, we also provide the Gaussian kernel used here.



\subsection{Main Results}
\paragraph{Qualitative Results}

We evaluate our method on both synthetic and real-world datasets, including MVImageNet~\cite{yu2023mvimgnet} and OmniObject3D~\cite{wu2023omniobject3d}. Selected cases are presented in Fig.~\ref{fig:demo} and more results can be found in the Appendix \ref{more result}. It can be seen that our approach enhances the smoothness of the generated mesh without compromising the texture quality. Additionally, the inference time is nearly the same as that of the base model, effectively providing a "free lunch" for 3D generation.

\paragraph{Quantitative Results}
Similar to previous works \cite{wang2024crm, xu2024instantmesh}, we evaluate the effectiveness of our approach on the Google Scanned Objects (GSO) \cite{downs2022google} dataset which is not included in the training dataset. We randomly select 30 shapes as our GSO test dataset. For each object, we render a $512\times 512$ size front-view image as the input image.

To evaluate different methods, we follow previous reconstruction works~\cite{mescheder2019occupancy, liu2020meshing, pumarola2022visco} to report Volume IoU, Chamfer Distance and Normal Consistency Score(NCS). 
Volume IoU and Chamfer distance describe the geometry quality between the generated mesh and ground-truth mesh, while the Normal Consistency Score tells the smoothness of generated meshes. Prior to testing, we carefully align the pose between the evaluated mesh and the ground truth mesh and scale them to fit within a $[-0.5, 0.5]^3$ box. The results are shown in table \ref{tab:quant1}.

\begin{table}[ht]

\centering
\caption{Quantitative comparison for the geometry quality. We report the metrics of Volume IoU, Chamfer Distance and NCS on GSO dataset.}
\begin{tabular}{@{}lcccc@{}}
\toprule
& \multicolumn{2}{c}{CRM~\cite{wang2024crm}} & \multicolumn{2}{c}{InstantMesh~\cite{xu2024instantmesh}} \\
\cmidrule(lr){2-3} \cmidrule(lr){4-5}
                    & base & w. Freeplane & base & w. Freeplane \\

\midrule

Vol. IoU.$\uparrow$    & 0.422 & \colorbox{red!10}{0.434} & 0.429 & \colorbox{red!30}{0.448} \\

Chamfer Dist.$\downarrow$($\times 10^{-3}$)    & \colorbox{red!10}{16.06} & 16.12 & 16.34 & \colorbox{red!30}{14.73} \\


NCS $\uparrow$($\%$)     & 67.01 & 67.30 & \colorbox{red!10}{68.20} & \colorbox{red!30}{69.37} \\

\bottomrule
\end{tabular}
\label{tab:quant1}
\end{table}



\subsection{Ablation Study}

\subsubsection{Frequency Filtering}
Based on our formulation, we adopt different frequency filtering strategies, including bilinear, GaussianBlur, and BilateralFilter.

\textbf{Bilinear}
Bilinear interpolation calculates new pixel values based on the weighted average of the four nearest pixels surrounding the desired location.

\textbf{Gaussian Blur} Gaussian blur is a widely used image blurring technique to reduce image noise and detail. It works by convoluting the image with a Gaussian kernel.

\textbf{Bilateral Filter} The bilateral filter is an edge-preserving and noise-reducing filter. It applies a weighted average of neighboring pixels to each pixel, considering both spatial distance and intensity difference.

\begin{figure}[tb] 
    \centering 

    \begin{minipage}[t]{0.3\linewidth} 
    \centering
        \includegraphics[height=5cm]{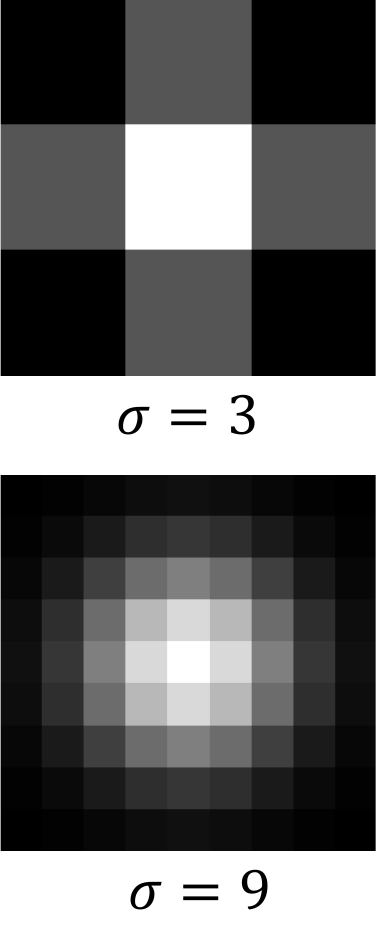} 
        \caption{Gaussian kernel.} 
        \label{fig:ablation_left} 
    \end{minipage}
    \hfill 
    \begin{minipage}[t]{0.68\linewidth} 
    \centering
        \includegraphics[height=5cm]{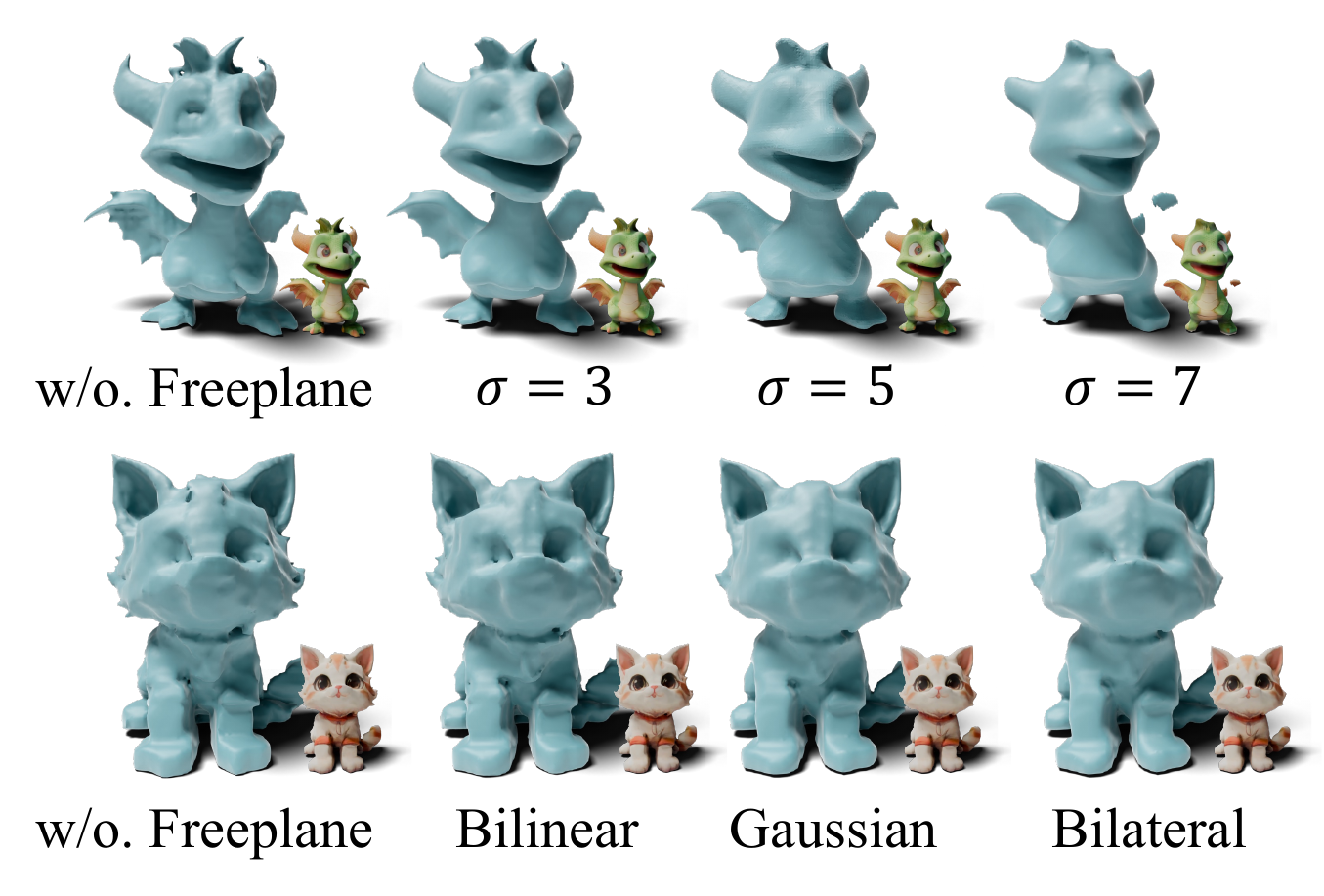} 
        \caption{The first row is from InstantMesh, using a bilateral filter with varying kernel sizes. The second row is from CRM with different filterings.} 
        \label{fig:ablation_right} 
    \end{minipage}

\end{figure}

As shown in Fig. \ref{fig:ablation_right}, in comparison to the original CRM, all these filters enhance the smoothness and geometric details of the generated meshes. Bilinear, being a fixed local averaging method, underperforms when the triplane resolution is high. Both Gaussian blur and bilateral filter produce similar effects, but the bilateral filter achieves superior local details. This is because the bilateral filter ensures local smoothing while preserving the triplane boundaries.










\subsubsection{Kernel Size}

We evaluate the effectiveness of our approach using various kernel sizes. As shown in Fig. \ref{fig:ablation_right}, larger kernels result in smoother surfaces, though excessively large kernels can cause a loss of geometric details. Overall, the kernel size is correlated to the resolution of triplanes. Specifically, higher resolution triplanes necessitate larger kernels to properly modulate high-frequency components.


\subsection{Limitations}    \label{limitation}
While our method enhances the geometric smoothness and 3D coherence of the generated meshes, some limitations remain unresolved.
First, our approach exhibits limited effectiveness in modeling slender structures, resulting in occasional omission of tiny details. Second, the choice of the frequency modulation strategy relies on the model architecture and the objects being generated, making it non-trivial to generalize. Integrating super-resolution and frequency modulation networks into the large reconstruction model could potentially enhance its capabilities, possibly requiring the retraining of feed-forward models from scratch. Due to the huge computation cost required, we leave this issue for the community to explore further. Nonetheless, we consider Freeplane as a simple yet effective baseline method to leverage the geometric priors of triplanes while maintaining robustness against inconsistent multi-view images.

\section{Conclusion}    \label{conclusion}

In this work, we present Freeplane, a simple yet effective approach that significantly enhances the generation quality of 3D feed-forward models without requiring any additional costs. Focusing on the primary challenge posed by inconsistent multi-view images, we analyze the key role of triplanes in feed-forward models. Our study reveals the correlation between the inconsistency of multi-view images and the local high-frequency artifacts in triplanes. Leveraging this insight, we suggest strategically filtering out high-frequency conflicts in the triplanes and combining triplanes before and after filtering to produce high-quality textured meshes. The artifact filtering operation effectively mitigates the negative effects caused by inconsistent multi-view images, leading to a smooth mesh surface while preserving geometric details. Our approach can be seamlessly integrated into existing triplane-based feed-forward models, enhancing their generation quality and robustness against inconsistent multi-view images. We hope that our research will inspire further exploration into constructing a training pipeline to exploit the frequency characteristics of triplanes.
\medskip

{
\small

\bibliographystyle{plain}
\bibliography{main}

}


\clearpage

\appendix

\section{Appendix}

\subsection{Metric Explanation}

As mentioned in the main paper, we follow ONet~\cite{mescheder2019occupancy} to report volume IoU, Chamfer distance and normal consistency score(NCS) to evaluate baselines and our method. To compute the normal consistency score, let $\mathcal{M}_{\text{pred}}$ and $\mathcal{M}_{\text{gt}}$ represent the sets of points inside or on the predicted mesh and ground truth mesh, respectively. The NCS can then be determined using the following formula:

\begin{equation}
    \text{NCS}(\mathcal{M}_{\text{pred}}, \mathcal{M}_{\text{gt}}) = \frac{1}{2} \left( \frac{1}{N} \sum_{i=1}^{N} \cos(\theta_{i}) + \frac{1}{N} \sum_{i=1}^{N} \cos(\phi_{i}) \right)
\end{equation}

Where \( \theta_{i} \) is the angle between the predicted normal vector and its nearest true normal vector, \( \phi_{i} \) is the angle between the true normal vector and its nearest predicted normal vector and \( \cos(\theta_{i}) \) and \( \cos(\phi_{i}) \) are the cosine similarities, computed as the dot product of the vectors normalized by their magnitudes.





\subsection{More Results}  \label{more result}

\begin{figure}[tb]
    \centering
    \includegraphics[width=\linewidth]{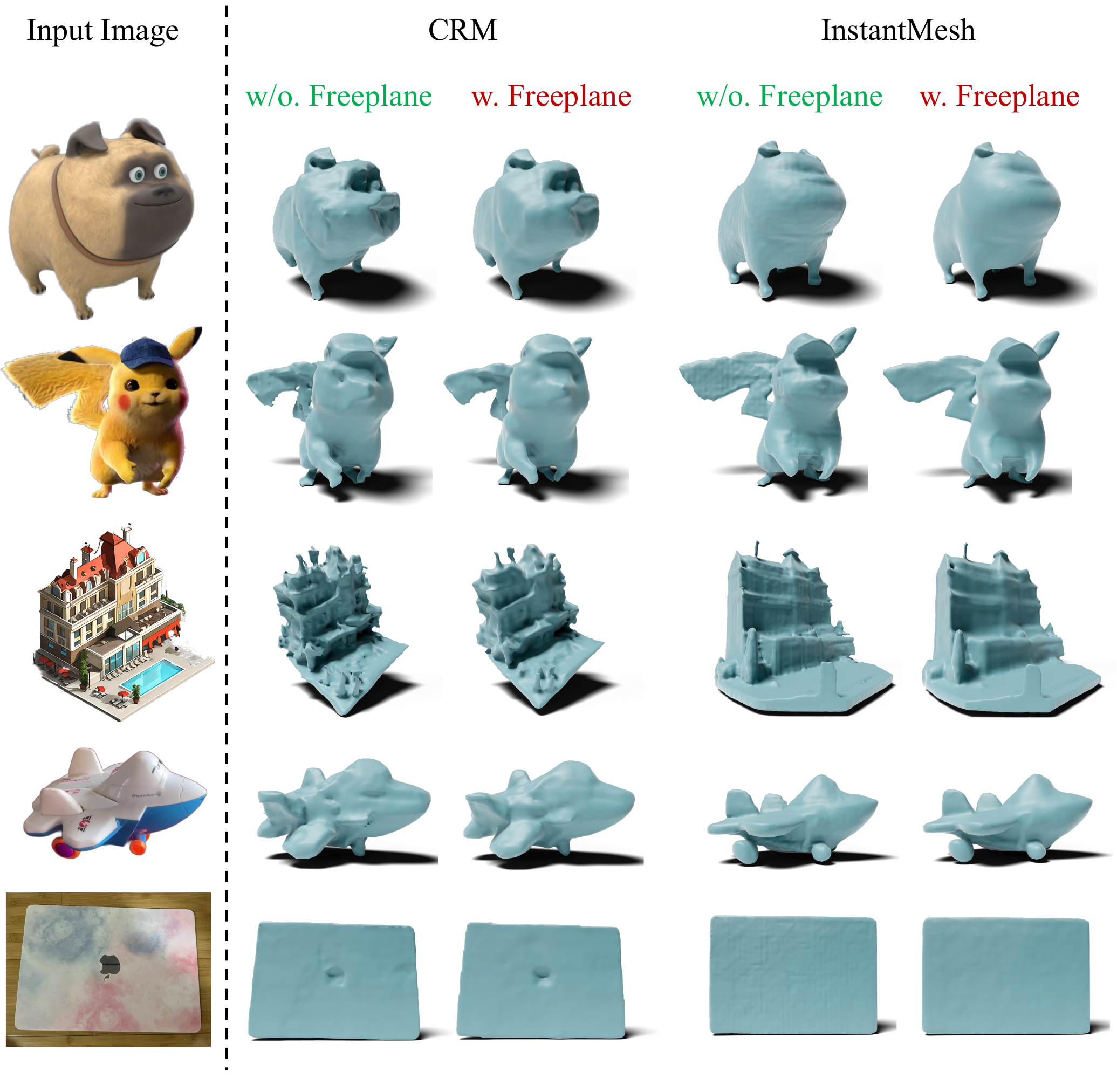}
    \caption{More results. Please zoom in to see the details of the meshes.}
    \label{fig:more}
\end{figure}

We provide additional qualitative results using images from the website, open-source real-world datasets, and our own photographed images. As shown in Fig.~\ref{fig:more}, our approach can be integrated into triplane-based sparse-view reconstruction models to improve the smoothness and geometric details.







\subsection{Pseudo Code}

We provide the pseudo code for our approach in Algorithm \ref{alg:code}. Our approach includes the frequency modulation and the combination of triplanes before and after filtering to generate textured meshes.

\begin{algorithm}[]

\caption{Freeplane  pseudo code.}\label{alg:code}
\newcommand{\hlbox}[1]{%
  \fboxsep=1.2pt\hspace*{-\fboxsep}\colorbox{blue!10}{\detokenize{#1}}%
}
\lstset{style=mocov3}
\vspace{-3pt}
\begin{lstlisting}[
    language=python,
    escapechar=@,
    label=code:freeplane]
    # Input list:
    # input multi-view images (img): [b, 6, c, h, w]
    # frequency_type: BilateralFilter
    # kernel_size: k
    # resolution: resolution
    # ctx: nvdiffrast.torch.RasterizeCudaContext()
    
    # Output:
    # 3D mesh: vertice, face, uvs, mesh_tex_idx, texture_map

    # Freeplane framework
    triplane = decoder(img)     # produce triplanes from multi-view images
    freeplane = BilateralFilter(triplane, kernel_size=k)        # filter the original triplanes
    vertice, face = get_geometry_prediction(freeplane)      # get_geometry_prediction from freeplane
    uvs, mesh_tex_idx, gb_pos, tex_hard_mask = xatlas_uvmap(ctx, vertice, face, resolution)   # get uv map
    texture = get_texture_prediction(triplane, gb_pos, tex_hard_mask)     # query the texture field
    background_feature = torch.zeros_like(texture)     # build background features
    img_feature = torch.lerp(background_feature, texture, tex_hard_mask)    # extract the texture
    texture_map = img_feature.permute(0, 3, 1, 2).squeeze(0)
    
    return vertice, face, uvs, mesh_tex_idx, texture_map
    
    
\end{lstlisting}\vspace{-5pt}
\end{algorithm}



\subsection{Multi-view Inconsistency in InstantMesh}   \label{motivation_instant}

\begin{figure}[tb]
    \centering
    \includegraphics[width=\linewidth]{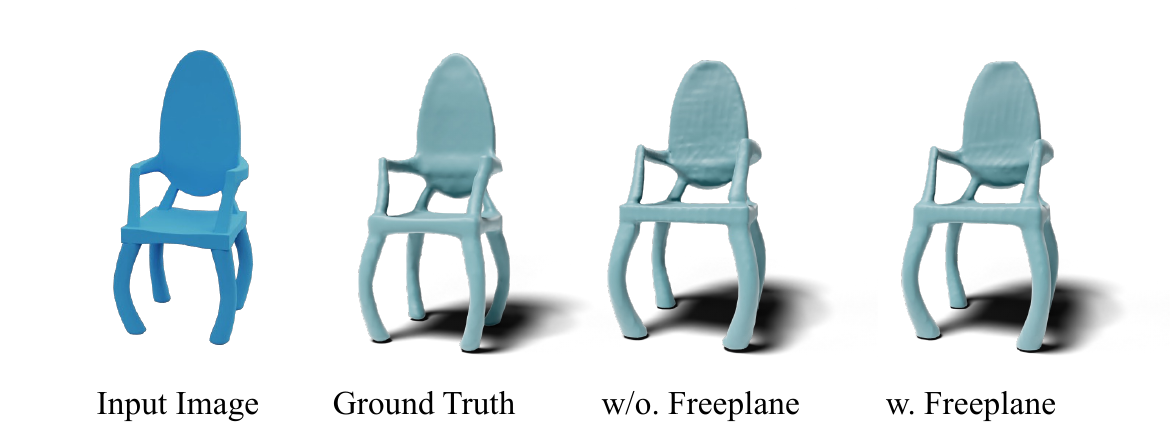}
    \caption{Visualization results for InstantMesh. Ground Truth means that ground-truth images are used to generate meshes, while w/o. Freeplane is the original InstantMesh, which leverages the multi-view images from Zero123++ as the inputs. Our approach Freeplane significantly improves the mesh quality.}
    \label{fig:GT_instantmesh}
\end{figure}

We provide the visualization results for InstantMesh. As shown in Fig. \ref{fig:GT_instantmesh}, the inconsistent multi-view images cause some apparent artifacts on the generated meshes. Our approach Freeplane substantially enhance the smoothness and geometric details in final meshes.

\subsection{Social Impact}   \label{impact}

As with many other generative models, our approach could be used to produce malicious 3D content, warranting additional caution.

\clearpage

\end{document}